\documentclass[10pt,twocolumn,letterpaper]{article}

 \usepackage{cvpr}              

\makeatletter

\definecolor{cvprblue}{rgb}{0.21,0.49,0.74}
\usepackage[pagebackref,breaklinks,colorlinks,allcolors=cvprblue]{hyperref}

\def\paperID{} 
\def\confName{cvpr}

\title{Learning from the Right Patches: A Two-Stage Wavelet-Driven Masked Autoencoder for Histopathology Representation Learning}

\author{Raneen Younis\\
PLRI Medical Informatics Institute\\
CAIMed Reserch Center\\
Hannover Medical School\\
{\tt\small Younis.Raneen@mh-hannover.de}
\and
Louay Hamdi\\
Computer Science Institute\\
Leibniz University Hannover\\
{\tt\small louay.hamdi@stud.uni-hannover.de}
\and
Lukas Chavez\\
Sanford Burnham Prebys Medical Discovery Institute\\
Rady Children’s Institute for Genomic Medicine\\
University of California San Diego\\
{\tt\small lchavez@sbpdiscovery.org}
\and
Zahra Ahmadi\\
PLRI Medical Informatics Institute\\
CAIMed Reserch Center\\
Hannover Medical School\\
{\tt\small Ahmadi.Zahra@mh-hannover.de}
}

\begin{document}
\maketitle

\begin{abstract}
Whole-slide images are central to digital pathology, yet their extreme size and scarce annotations make self-supervised learning essential.
Masked Autoencoders (MAEs) with Vision Transformer backbones have recently shown strong potential for histopathology representation learning.
However, conventional random patch sampling during MAE pretraining often includes irrelevant or noisy regions, limiting the model’s ability to capture meaningful tissue patterns.
In this paper, we present 
, a lightweight and domain-adapted framework that brings structure and biological relevance into MAE-based learning through a wavelet-informed patch selection strategy. WISE-MAE applies a two-step coarse-to-fine process: wavelet-based screening at low magnification to locate structurally rich regions, followed by high-resolution extraction for detailed modeling. This approach mirrors the diagnostic workflow of pathologists and improves the quality of learned representations. Evaluations across multiple cancer datasets, including lung, renal, and colorectal tissues, show that WISE-MAE achieves competitive representation quality and downstream classification performance while maintaining efficiency under weak supervision.
\end{abstract}    
\section{Introduction}
Histopathology remains one of the most essential tools in cancer diagnosis, providing fine-grained insight into tissue morphology that directly influences clinical decision-making. With the transition from glass slides to digitized whole-slide images (WSIs), computational pathology has emerged as a powerful area for applying machine learning to histological analysis. WSIs capture vast amounts of visual information—often exceeding $100,000$ pixels per dimension—making them rich in diagnostic content but also computationally challenging to process. To manage this complexity, most approaches divide WSIs into smaller image patches, often using the Multiple Instance Learning (MIL) framework~\cite{18}, where a slide is treated as a bag of patches.
\begin{figure}[t]
\centering
\includegraphics[width=\columnwidth]{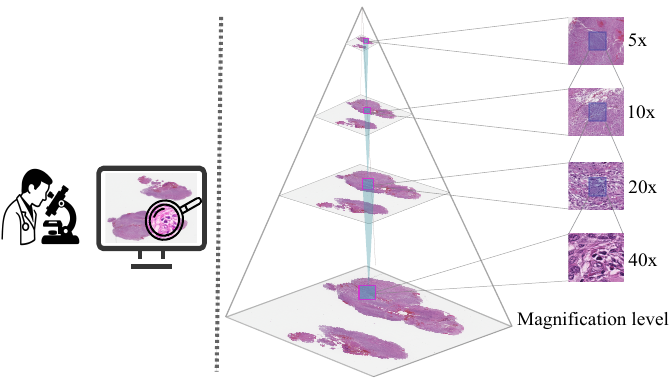}
\caption{Illustration of hierarchical analysis in computational pathology. Left: A pathologist examines a whole-slide image and zooms into regions of diagnostic interest, a process mimicked in computational models. Right: Multi-resolution representation of the same WSI across magnification levels. Lower layers correspond to higher magnification with finer tissue detail, reflecting the increasing structural information accessed through zooming.}
\label{fig1}
\end{figure}
While MIL methods have demonstrated strong performance in WSI classification, they face several limitations in practice. First, not all extracted patches contribute equally to the diagnostic task. Random sampling often captures large portions of background, fat, or benign tissue~\cite{10,12}, which reduces the overall effectiveness of downstream task learning. In addition, many MIL-based models assume independence among patches~\cite{zhang2022dtfd, kanavati2020weakly}, thereby ignoring the spatial and morphological continuity that exists across tissue structures. To address this, recent approaches have adopted transformers to model inter-patch dependencies more effectively~\cite{shao2021transmil}. However, a more fundamental challenge persists: the sparsity of relevant diagnostic regions. Tumor prevalence within whole-slide images varies significantly across tasks. In some cases, tumor regions occupy only a small fraction of the slide (e.g., metastasis detection), making informative patch identification challenging, whereas in others (e.g., primary tumor slides), tumor coverage is more extensive. Without explicit guidance, random sampling often includes background or benign regions, diluting the model's focus on diagnostically relevant features.

Recent advances in self-supervised learning (SSL) have opened new possibilities for medical imaging, where labeled data is often limited. In particular, masked image modeling (MIM) techniques such as masked autoencoders \cite{15} have shown strong potential by learning to reconstruct hidden parts of images without requiring manual annotations. Compared to contrastive learning \cite{7,14}, MAEs offer a more computationally friendly training pipeline, free from the need for large batch sizes or carefully curated positive/negative pairs, making them especially suitable for pathology, where access to GPUs and annotated data is often constrained~\cite{2,26}.
However, the effectiveness of MAEs in histopathology still depends on the quality of the patches used during training. Many existing approaches rely on random patch sampling, which can overwhelm the model with irrelevant or low-content regions~\cite{10,quan2024global}. Moreover, most foundational models assume access to huge datasets and high computational budgets, which are not always feasible in real-world clinical settings.

To address these challenges, we propose \underline{W}avelet-\underline{I}nformed \underline{S}ampling for \underline{E}ncoding \underline{M}asked \underline{A}uto\underline{E}ncoder (\textbf{WISE-MAE}), a lightweight, two-stage self-supervised framework tailored for histopathology. The design of WISE-MAE is inspired by how pathologists navigate WSIs, as illustrated in Figure~\ref{fig1}. Typically, a pathologist first scans the slide at low magnification to locate suspicious regions and then zooms in to inspect fine-grained tissue details. Mimicking this diagnostic process, our method first performs wavelet-based frequency analysis at low resolution to identify the most informative tissue areas. From these selected regions, we extract high-resolution patches and apply masked autoencoding to encourage the model to learn both detailed intra-patch features and broader contextual patterns.
This two-stage training strategy allows WISE-MAE to maximize data efficiency while remaining computationally lightweight, making it practical for setups with limited hardware resources. 
Our approach aligns naturally with the hierarchical and multi-scale structure of tissue morphology and supports efficient pretraining under weak supervision using only slide-level labels. 
In summary, our main contributions are as follows:

\begin{itemize}
\item \textbf{Wavelet-guided Multi-resolution Patch Selection:} We propose a biologically inspired patch selection strategy that mimics the workflow of pathologists. Specifically, we use wavelet-based frequency scoring at intermediate resolution to identify structurally rich tissue regions, from which high-resolution patches are extracted for downstream processing. This component constitutes the core novelty of our framework and provides a principled way to guide representation learning toward biologically meaningful content.

\item \textbf{MAE Adaptation for Histopathology:} We adapt existing Vision Transformer-based Masked Autoencoder (ViT-MAE) frameworks to histopathology by integrating our proposed sampling strategy, enabling the model to learn meaningful representations from high-resolution tissue patches in a self-supervised manner.

\item \textbf{Label-efficient Focus on Informative Regions:} By combining wavelet-based scoring with multi-resolution patch selection, our method guides the model's attention toward diagnostically relevant areas of WSIs, without the need for pixel-level labels. This makes our framework especially suitable for weakly supervised and resource-constrained settings.

\item \textbf{Comprehensive Empirical Evaluation:} We evaluate WISE-MAE on multiple downstream classification tasks using Camelyon16 \cite{bejnordi2017diagnostic}, TCGA-RCC, and TCGA-NSCLC \cite{tomczak2015review} datasets. Experimental results demonstrate that our approach consistently outperforms baseline methods and achieves competitive performance.
\end{itemize}

\section{Related Work}
\subsection{Multiple Instance Learning}
Bag-level classification methods in computational pathology typically fall into two broad categories: bag-based and instance-based approaches. Bag-based methods attempt to train instance classifiers by assigning pseudo-labels to individual patches and then aggregating the predictions of the top-$k$ instances to make a final slide-level (bag-level) prediction~\cite{chikontwe2020multiple, xu2019camel,hou2016patch,4,kanavati2020weakly}. In contrast, instance-based methods focus on learning meaningful representations from individual instances (patches) and then aggregate these features to form a high-level bag representation, which is directly used to predict the slide label~\cite{sharma2021cluster,shao2021transmil, lu2021data,zhang2022dtfd,19}. Empirical evidence suggests that instance-based methods often outperform their bag-based counterparts in real-world histopathology tasks.
One widely adopted instance-based method is attention-based multiple instance learning (ABMIL) \cite{19}, which proposes computing attention weights for each instance using a trainable neural attention mechanism. TransMIL \cite{shao2021transmil} takes a different approach by employing transformer \cite{NIPS2017_3f5ee243} architectures to explicitly model interactions among instances in a bag, enabling it to capture spatial and contextual dependencies across a whole slide image
However, most MIL pipelines still rely on random patch sampling, potentially including large non-informative background regions that can harm learning performance.
\begin{figure*}[t]
\centering
\includegraphics[width=0.96\textwidth]{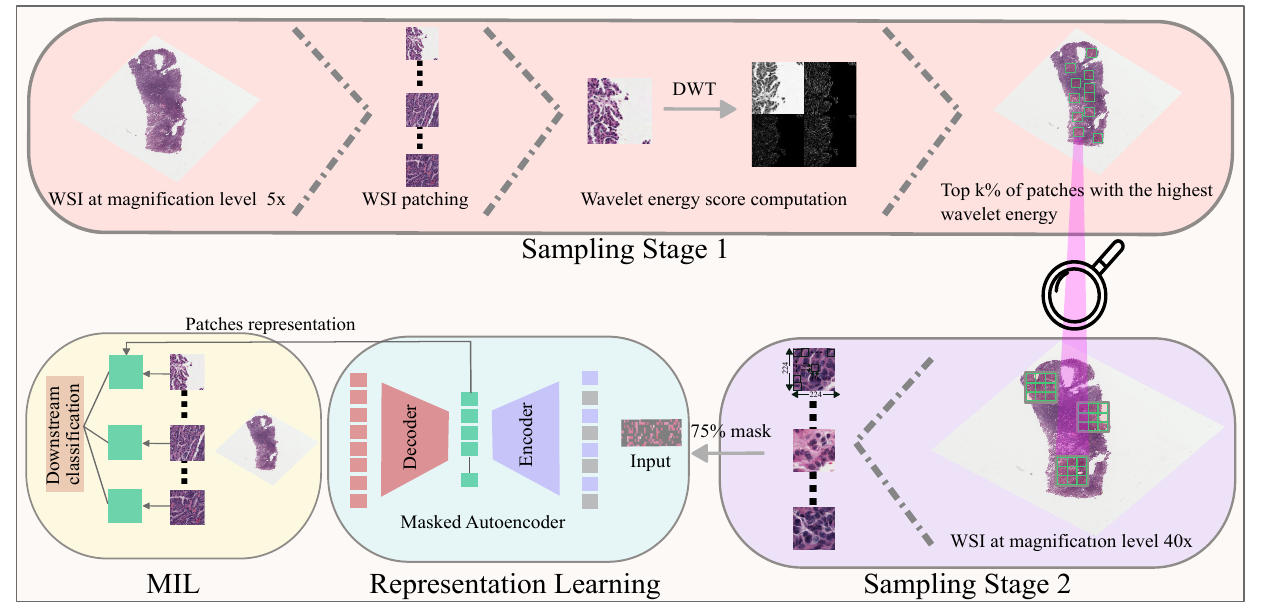}
\caption{Overview of the WISE-MAE framework. The process begins with Stage 1 patch sampling at $5$× magnification (low resolution), where wavelet energy is computed for each patch to assess morphological richness. The top-$k$\% patches with the highest energy scores are selected. In Stage 2, these selected regions are revisited at $40$× magnification (high resolution) to extract fine-grained patches containing rich tissue detail. These high-resolution patches are then used to train a Masked Autoencoder (MAE) in a self-supervised fashion. The learned encoder representations are subsequently used in an MIL framework for downstream classification tasks.}
\label{fig2}
\end{figure*}
\subsection{Self-Supervised Learning in Pathology}

Self-supervised learning (SSL) has become a promising approach for histopathology, enabling models to learn transferable representations from unlabeled whole-slide images \cite{chowdhury2021applying,boyd2021self}. While contrastive methods have shown early success by adapting techniques from natural images~\cite{shurrab2022self,zhang2022contrastive}, their performance often depends on carefully crafted augmentations~\cite{reed2021selfaugment,xu2020data,yang2021self} or domain-specific pretext tasks~\cite{li2021sslp}. However, the visual homogeneity of histology slides and reliance on cropped views can limit their generalization ability.
Masked autoencoders (MAEs) ~\cite{2}, part of the Masked Image Modeling (MIM) paradigm, have recently emerged as a compelling alternative~\cite{wei2022masked}, offering a more efficient and augmentation-free training pipeline. While Global Contrast-Masked Autoencoders~\cite{quan2024global} apply masking at the WSI grid level to emphasize global contextual learning, our approach focuses on frequency-based patch selection to prioritize structurally informative regions before masked reconstruction.

\subsection{Leveraging Multi-Resolution Information in Histopathology}
Pathologists rely on multiple magnification levels when examining tissue, making multi-resolution analysis essential for WSI-based diagnosis~\cite{tokunaga2019adaptive, zhou2022deep}. Prior studies typically either combine features from different resolutions~\cite{li2021multi} or treat them as separate inputs within the same bag~\cite{hashimoto2020multi}. Closest to our work is Hierarchical Attention-Guided
Multiple Instance Learning (HAG-MIL)~\cite{xiong2023diagnose}, which explores pyramidal masking across resolutions. Unlike these approaches, our method integrates multi-resolution information hierarchically by first identifying informative regions at low magnification via wavelet analysis, then focusing masked autoencoding on high-resolution patches, in a way that mirrors how pathologists examine slides.


\subsection{Patch Sampling Strategies}
Existing patch selection strategies can be broadly categorized into random, attention-guided, and clustering-based approaches. Random sampling is simple and scalable but often captures low-content regions with limited diagnostic value. Attention-guided methods, such as those used in ABMIL~\cite{19}, prioritize patches based on learned relevance scores, while clustering-based frameworks~\cite{chenni2019patch} promote diversity by selecting representative instances from feature-space groupings. Frequency-domain representations have also been explored to enhance masked image modeling, as in FreMIM~\cite{wang2024fremim}, which incorporates Fourier features for medical image reconstruction. In contrast, our method leverages wavelet transforms that provide localized, multi-scale frequency decomposition, enabling biologically grounded and fully unsupervised patch selection. By quantifying wavelet energy, WISE-MAE focuses on structurally rich tissue regions, improving the efficiency of self-supervised pretraining on histopathology slides.

\section{The WISE-MAE Framework}
The WISE-MAE framework introduces a wavelet-guided two-stage sampling pipeline to enhance masked autoencoder pretraining for whole-slide image representation learning under weak supervision. The overall workflow, as illustrated in Figure~\ref{fig2}, consists of four main stages: low-resolution patch sampling with wavelet scoring, high-resolution patch refinement, MAE-based self-supervised pre-training, and MIL-based downstream classification.

\subsection{Stage 1: Low-Resolution Patch Sampling via Wavelet Energy}
Given the extreme size of WSIs, we begin by extracting non-overlapping image patches at a coarse magnification level (e.g. $10$x). These low-resolution patches serve as the first filter for content-aware sampling. To quantify morphological richness, we apply a single-level 2D Discrete Wavelet Transform (DWT) to each patch using the Daubechies basis. This decomposes the image into four subbands: $LL$ (low-frequency approximation) and $LH$, $HL$, $HH$ (high-frequency detail coefficients in horizontal, vertical, and diagonal directions). The wavelet energy is computed as the sum of squared coefficients from the high-frequency bands:
\begin{equation}
E(x) = \sum_{i,j} \left( LH(i,j)^2 + HL(i,j)^2 + HH(i,j)^2 \right).
\end{equation}
The wavelet energy criterion was chosen because high-frequency detail coefficients correspond to edge-rich and texture-dense regions, which typically represent diagnostically relevant morphological features such as cellular boundaries, nuclei clusters, and stromal textures. We rank all patches by their wavelet energy scores and retain the top $k\%$ (typically $20$\%) for further inspection.

\subsection{Stage 2: High-Resolution Patch Refinement}
After identifying the top $k\%$ wavelet-rich regions from Stage 1, we proceed to extract the corresponding high-resolution patches at $40\times$ magnification using the original slide coordinates. This step captures more granular histological structures such as nuclei, mitotic figures, and glandular architecture, which are not visible at lower magnifications. Each selected low-resolution patch center is mapped back to its high-resolution coordinates, from which we extract a $224\times224$ pixel patch. Although wavelet energy is computed independently for each patch, the spatial topology of the WSI is preserved through coordinate indexing, ensuring that the transition from low- to high-resolution sampling maintains the original tissue layout.

This two-stage, coarse-to-fine pipeline avoids redundant sampling, focuses computational resources on informative tissue areas, and mimics the diagnostic approach of human pathologists, who zoom in on suspicious areas after initial low-magnification screening.

\subsection{Self-Supervised MAE Pretraining}
\label{sec:mae_architecture}
The final set of high-resolution patches is used to train a Masked Autoencoder (MAE)~\cite{2}. We adopt the MAE with a ViT-Base encoder, which provides a strong trade-off between representational power and training feasibility. Smaller ViT variants, such as ViT-Small or ViT-Tiny, often lack the capacity to model the rich textural and spatial complexity of histopathology. In contrast, ViT-Base enables expressive learning while being computationally practical, as demonstrated in prior medical imaging benchmarks~\cite{9}.

\subsubsection*{Encoder Design}
The encoder \( f_{\text{enc}} \) follows the ViT-Base configuration with 12 transformer blocks, 12 attention heads, and a hidden embedding dimension of \( D = 768 \). 
Each input tile of size \( 224 \times 224 \) is divided into \( N = 196 \) non-overlapping patches of size \( 16 \times 16 \).
Each patch is flattened into a vector \( x_i \in \mathbb{R}^{P} \), where \( P = 16 \times 16 \times 3 = 768 \), 
and linearly projected into the embedding space using:
\begin{equation}
z_i = W_e \cdot \text{flatten}(x_i) + p_i, \quad W_e \in \mathbb{R}^{D \times P}, \quad p_i \in \mathbb{R}^{D}.
\end{equation}
We adopt the MAE masking strategy, where 75\% of the patches are randomly masked and omitted from the encoder input. 
Only the visible subset \( \mathbf{X}_{\text{vis}} \in \mathbb{R}^{V \times P} \), with \( V = 49 \), is processed by the encoder:
\begin{equation}
\mathbf{Z}_{\text{vis}} = f_{\text{enc}}(\mathbf{X}_{\text{vis}}) \in \mathbb{R}^{V \times D}.
\end{equation}
This approach enforces learning from sparse context without direct access to masked content.

\subsubsection*{Decoder Design}
The decoder \( f_{\text{dec}} \) is a lightweight transformer with 4 layers and an embedding dimension of 512. 
Encoder outputs are first projected to match this dimension through a linear layer. 
The decoder reconstructs pixel values for the masked patches using the encoded visible tokens, 
learnable mask tokens, and shared positional encodings for all \( N = 196 \) positions. 
Masked tokens are added only at this stage, consistent with the MAE formulation. 
Reconstruction is performed in RGB pixel space, and the decoder is discarded after pretraining.

This asymmetric encoder–decoder design promotes semantically meaningful encoder representations 
while maintaining computational efficiency. 
Note that “multi-resolution” in our framework refers to the hierarchical patch sampling stage, 
not to reconstruction, which operates at a single (40$\times$) scale.

\subsubsection*{Training Objective}
The reconstruction objective is the mean squared error (MSE) between predicted and true pixel values of masked patches:
\begin{equation}
\mathcal{L}_{\text{MSE}} = \frac{1}{|M|} \sum_{i \in M} \| \hat{x}_i - x_i \|^2,
\end{equation}
where \( M \) is the set of masked patch indices, \( \hat{x}_i \) is the predicted output, and \( x_i \) is the ground-truth pixel vector. This formulation ensures the encoder learns meaningful latent representations from incomplete visual contexts.

For the contrastive variant, an InfoNCE loss is incorporated to enhance representation discrimination:
\begin{equation}
\mathcal{L}_{\text{con}} = -\log 
\frac{\exp(\text{sim}(z_i, z_i^+)/\tau)}{\sum_{j=1}^{N}\exp(\text{sim}(z_i, z_j^-)/\tau)},
\end{equation}
where \(\text{sim}(\cdot,\cdot)\) denotes cosine similarity and \(\tau\) is the temperature parameter. The total objective combines both components as
\begin{equation}
\mathcal{L} = \mathcal{L}_{\text{MSE}} + \lambda\,\mathcal{L}_{\text{con}}.
\end{equation}

After training, the decoder is removed, and the pretrained encoder is retained to extract patch-level embeddings for downstream tasks.


\subsection{Downstream Classification via MIL}
After self-supervised pretraining, the encoder is frozen and used to extract patch-level feature embeddings from high-resolution WSI tiles. These embeddings represent semantically rich tissue characteristics learned during masked autoencoding and serve as input for downstream classification tasks.

Under a weakly supervised multiple instance learning (MIL) paradigm, each whole-slide image is treated as a bag of instance embeddings, with supervision provided only at the slide level. Various MIL architectures can be employed in this setting, including attention-based pooling \cite{19}. 

In this work, we adopt an attention-based MIL formulation using the CLAM framework~\cite{lu2021data}, which learns to assign soft importance weights to each patch and aggregates features via a gated attention mechanism. The final slide-level prediction is obtained through a fully connected layer applied to the weighted feature representation. During training, only the parameters of the aggregation and classification modules are updated, while the pretrained encoder remains fixed.

\begin{table*}[tb]
\centering
\begin{tabular}{lccc|ccc|ccc}
\toprule
\textbf{Model} & 
\multicolumn{3}{c}{\textbf{TCGA-NSCLC}} & 
\multicolumn{3}{c}{\textbf{TCGA-RCC}} & 
\multicolumn{3}{c}{\textbf{CAMELYON16}} \\
\cmidrule(lr){2-4} \cmidrule(lr){5-7} \cmidrule(lr){8-10}
& Acc & AUC & F1 
& Acc & AUC & F1 
& Acc & AUC & F1 \\
\midrule
Max Pooling & 0.774 & 0.863 & 0.774 & 0.880 & 0.970 & 0.847 & 0.682 & 0.715 & 0.549 \\
AB-MIL \cite{19} & 0.817 & 0.903 & 0.817 & 0.892 & 0.978 & 0.864 & 0.868 & 0.903 & 0.850 \\
CLAM-SB \cite{lu2021data} & 0.824 & 0.905 & 0.824 & 0.878 & 0.970 & 0.846 & 0.819 & 0.834 & 0.782 \\
CLAM-MB \cite{lu2021data} & 0.818 & 0.900 & 0.818 & 0.895 & 0.975 & 0.871 & 0.826 & 0.876 & 0.800 \\
TransMIL \cite{shao2021transmil} & 0.825 & 0.902 & 0.825 & 0.876 & 0.972 & 0.850 & 0.809 & 0.838 & 0.787 \\
DTFD-MIL \cite{zhang2022dtfd} & 0.838 & 0.902 & 0.830 & 0.898 & 0.976 & 0.884 & 0.899 & 0.933 & 0.858 \\
MAE \cite{2} & 0.841 & 0.910 & 0.841 & 0.885 & 0.962 & 0.860 & 0.826 & 0.856 & 0.748 \\
HAG-MIL \cite{xiong2023diagnose} & 0.849 & 0.921 & 0.849 & 0.914 & 0.982 & 0.894 & 0.887 & 0.946 & 0.874 \\
DS-MIL \cite{li2021dual} & \textbf{0.888} & 0.939&0.876 & 0.929 &0.984& 0.890 & 0.856 & 0.899 & 0.815 \\
DTFD-MIL \cite{zhang2022dtfd} & 0.824 & 0.887 & 0.823 & 0.962 & 0.991 & 0.962 & 0.897 & 0.945 & 0.864 \\
GCMAE \cite{quan2024global} & 0.857 & 0.925 & 0.857 & 0.905 & 0.979 & 0.889 & 0.864 & 0.909 & 0.829 \\
WiKG \cite{li2024dynamic} & 0.840 & 0.907 & 0.839 & \textbf{0.970} & \textbf{0.996} & \textbf{0.970} & 0.900 & 0.925 & 0.883 \\
\midrule
WISE-MAE & 0.873 & 0.933 & 0.873 & 0.898 & 0.974 & 0.872 & 0.880 & 0.918 & 0.836 \\
WISE-MAE+Contrastive & 0.887 & \textbf{0.944} & \textbf{0.887} & 0.917 & 0.985 & 0.896 & \textbf{0.905} & \textbf{0.953} & \textbf{0.889} \\
\bottomrule
\end{tabular}
\caption{Performance comparison across TCGA-NSCLC, TCGA-RCC, and CAMELYON16. Metrics include Acc, AUC, and F1-score. The best results for each dataset are highlighted in bold.}
\label{table:all_results}
\end{table*}

\section{Experimental Analysis}
\subsection{Datasets and Metrics}
We evaluate our approach using several publicly available whole-slide image datasets covering a range of organs and cancer types. All datasets provide slide-level labels, making them suitable for weakly supervised learning under a multiple instance learning (MIL) framework.
\begin{itemize}[nosep,leftmargin=*]
    \item \textbf{TCGA-NSCLC:} The Cancer Genome Atlas Non-Small Cell Lung Cancer (TCGA-NSCLC) dataset \cite{tomczak2015review} comprises two major subtypes: lung adenocarcinoma (LUAD) and lung squamous cell carcinoma (LUSC). A total of 993 diagnostic WSIs were included, with 507 LUAD and 486 LUSC slides. Patch extraction at 40$\times$ magnification results in approximately 9,958 patches per slide on average. The data is split into training, validation, and test sets using a 7:1:2 ratio.
    \item \textbf{TCGA-RCC:} The TCGA Renal Cell Carcinoma (TCGA-RCC) cohort \cite{tomczak2015review} includes 884 WSIs from three renal cancer subtypes: clear cell (KIRC), papillary (KIRP), and chromophobe (KICH). The slides yield an average of 13,907 high-resolution patches at 40$\times$ magnification. Data is divided into training, validation, and test subsets using the same 7:1:2 split.
    \item \textbf{CAMELYON16:} CAMELYON16 is a benchmark dataset \cite{bejnordi2017diagnostic} for lymph node metastasis detection in breast cancer. It contains 400 WSIs annotated with binary slide-level labels (metastatic vs. normal). All slides are scanned at 40$\times$ magnification, reflecting realistic clinical variability in staining and preparation.
\end{itemize}
To assess model performance, we report three standard classification metrics: Accuracy (Acc), F1-score, and area under the ROC curve (AUC). These metrics provide a comprehensive view of both overall and class-specific prediction quality. 

\subsection{Implementation Details}
For MAE pretraining, we use approximately 80\% of slides from TCGA-NSCLC and TCGA-RCC, as well as CAMELYON16, which are selected to ensure diverse tissue coverage. Patches are sampled via wavelet-guided selection at 10$\times$ and extracted at 40$\times$ resolution. Pretraining is performed using ViT-Base with a patch size of $16 \times 16$, embedding dimension 768, and 75\% random masking. The model is trained for 300 epochs using AdamW with a learning rate of $(\text{lr} = 1 \times 10^{-4}$, weight decay of 0.05, and cosine decay schedule. For downstream evaluation, the remaining 20\% of slides are used across binary and multiclass classification tasks. Patch embeddings are extracted from the frozen encoder and passed to a CLAM-based attention MIL classifier. All downstream models are trained with Adam $(\text{lr} = 2 \times 10^{-4}, \text{weight decay} = 1 \times 10^{-5})$, batch size 1, and early stopping on validation loss. Features and metadata are stored in HDF5 format. All experiments are run on 4 NVIDIA H100 GPUs using PyTorch.

\subsection{Comparison with State-of-the-arts}
We compare the performance of WISE-MAE with state-of-the-art MIL and self-supervised frameworks on the TCGA-NSCLC, TCGA-RCC, and CAMELYON16 datasets (Table~\ref{table:all_results}). Unlike most existing WSI classification methods, WISE-MAE is engineered for efficiency, containing approximately 100K trainable parameters—substantially fewer than typical attention-based MIL models, which are reported to have between 0.8M and 1.2M parameters in recent MIL benchmarks~\cite{keshvarikhojasteh2024multi}.

Most existing MIL approaches rely on either random or attention-driven patch sampling. For example, Max Pooling and MAE~\cite{2} employ random sampling without prioritization, whereas ABMIL~\cite{19}, CLAM~\cite{lu2021data}, TransMIL~\cite{shao2021transmil}, and DS-MIL~\cite{li2021dual} utilize learned attention mechanisms to identify diagnostically relevant regions. Clustering-based methods such as CLAM-SB~\cite{lu2021data} and DTFD-MIL~\cite{zhang2022dtfd} further increase complexity by combining attention pooling with instance-level clustering. GCMAE~\cite{quan2024global} enhances MAE pretraining through global context modeling but still depends on random patch selection.

In contrast, WISE-MAE introduces a biologically motivated, fully unsupervised wavelet-guided patch selection strategy that prioritizes structurally informative tissue regions before pretraining, thereby reducing redundancy and enhancing data utilization. Despite its compact architecture, WISE-MAE achieves comparable or superior results on two of the three benchmark datasets (TCGA-NSCLC and CAMELYON16), demonstrating that targeted patch selection can deliver competitive performance even with a lightweight model. Furthermore, the contrastive variant of WISE-MAE improves discriminative capability without increasing model depth or parameter count.



\subsection{Generalizability of WISE-MAE}
\label{sec:wise_mae_generalization}
To evaluate the generalization capability of WISE-MAE, we conducted three transfer experiments spanning inter-organ, cross-domain, and cross-task settings: (i) colorectal tissue classification after pretraining on lung cancer slides, (ii) lung cancer subtyping after pretraining on renal cancer, and (iii) metastasis detection after pretraining on renal cancer.

\noindent\textbf{TCGA-NSCLC \textrightarrow{} NCT-CRC:}
This setting assesses robustness to significant tissue type and staining variations. As shown in Table~\ref{tab:nsclc_to_nctcrc}, WISE-MAE and its contrastive variant consistently outperform ViT and ResNet baselines, including both random and ImageNet-initialized versions. While standard MAE and GCMAE models already show notable improvements over baselines, the WISE-MAE variants achieve further gains, with the contrastive version reaching the highest accuracy (0.937) and AUC (0.979). These results highlight the benefit of spatially aware, multi-resolution patch selection, and demonstrate that the contrastive objective strengthens representation consistency across tissue domains, enhancing transferability across tissue types.

\noindent\textbf{TCGA-RCC \textrightarrow{} TCGA-NSCLC:}
This experiment tests cross-organ generalization by transferring from renal cancer classification to lung cancer subtyping. As shown in Table~\ref{tab:rcc_to_nsclc}, WISE-MAE and its contrastive variant outperform all baseline models, including both random and ImageNet-initialized ResNet50 and ViT architectures. While MAE already provides a strong baseline, WISE-MAE achieves notable improvements in AUC (0.982 vs. 0.961) and F1-score (0.940 vs. 0.910), confirming the benefit of structured patch selection. The contrastive variant further boosts performance across all metrics, indicating that the contrastive objective contributes to more transferable and discriminative feature representations across organ types.

\begin{table}[t]
\centering
\begin{tabular}{lccc}
\toprule
\textbf{Model} & \textbf{Acc} & \textbf{AUC} & \textbf{F1}  \\
\midrule
Random ViT-B/16           & 0.684 & 0.742 & 0.660  \\
ImageNet ViT-B/16         & 0.736 & 0.790 & 0.710  \\
Random ResNet50           & 0.702 & 0.768 & 0.690  \\
ImageNet ResNet50         & 0.751 & 0.825 & 0.730 \\
MAE                       & 0.876 & 0.931 & 0.880  \\
GCMAE                     & 0.891 & 0.945 & 0.890  \\
WISE-MAE                  & 0.908 & 0.962 & 0.910  \\
WISE-MAE + Contrastive    & \textbf{0.923} & \textbf{0.972} & \textbf{0.930}  \\
\bottomrule
\end{tabular}
\caption{Cross-Task Transfer from TCGA-RCC (Renal Cancer) to CAMELYON16 (Lymph Node Metastasis Detection). Evaluation of generalization performance across WISE-MAE and baseline methods under organ and diagnostic domain shift.}
\label{tab:rcc_to_camelyon}
\end{table}

\noindent\textbf{TCGA-RCC \textrightarrow{} CAMELYON16:}
This task represents a more difficult transfer setting involving both organ and diagnostic domain shift—from renal cancer classification to lymph node metastasis detection. As reported in Table~\ref{tab:rcc_to_camelyon}, WISE-MAE again achieves strong results, outperforming MAE, GCMAE, and traditional supervised baselines across all metrics. The contrastive WISE-MAE variant yields the highest AUC (0.972) and F1-score (0.930), indicating its robustness and improved feature generalization under substantial cross-task and cross-domain shifts.

\begin{table}[t]
\centering
\begin{tabular}{lccc}
\toprule
\textbf{Model} & \textbf{Acc} & \textbf{AUC} & \textbf{F1}  \\
\midrule
Random ViT-B/16           & 0.721 & 0.788  & 0.700 \\
ImageNet ViT-B/16         & 0.774 & 0.842  & 0.760  \\
Random ResNet50           & 0.741 & 0.812  & 0.730  \\
ImageNet ResNet50         & 0.793 & 0.861  & 0.780 \\
MAE                       & 0.868 & 0.935 & 0.870  \\
GCMAE                     & 0.881 & 0.951 & 0.890  \\
WISE-MAE                  & 0.906 & 0.961 & 0.900 \\
WISE-MAE + Contrastive    & \textbf{0.937} & \textbf{0.979} & \textbf{0.930}  \\
\bottomrule
\end{tabular}
\caption{Cross-Domain Transfer from TCGA-NSCLC (Lung Cancer) to NCT-CRC (Colorectal Tissue Classification). Evaluation of WISE-MAE and its contrastive variant under substantial tissue type and staining variations.}
\label{tab:nsclc_to_nctcrc}
\end{table}
\begin{table}[t]
\centering
\begin{tabular}{lccc}
\toprule
\textbf{Model} & \textbf{Acc} & \textbf{AUC} & \textbf{F1} \\
\midrule
Random ViT-B/16  & 0.762 & 0.893 & 0.770  \\
ImageNet ViT-B/16 & 0.825 & 0.918 & 0.810  \\
Random ResNet50 & 0.864 & 0.947 & 0.850  \\
ImageNet ResNet50 & 0.886 & 0.961 & 0.870  \\
MAE & 0.891 & 0.933 & 0.910  \\
WISE-MAE & 0.910 & 0.979 & 0.930  \\
WISE-MAE + Contrastive & \textbf{0.926} & \textbf{0.982} & \textbf{0.940}  \\
\bottomrule
\end{tabular}
\caption{Cross-Organ Transfer from TCGA-RCC (Renal Cancer) to TCGA-NSCLC (Lung Cancer Subtyping). Performance comparison of WISE-MAE and baseline methods, highlighting improvements in AUC and F1 through structured patch selection and contrastive pretraining.}

\label{tab:rcc_to_nsclc}
\end{table}

The improved transferability of WISE-MAE can be attributed to its focus on morphology-driven rather than appearance-based representations. The wavelet-guided sampling emphasizes patches with rich structural and textural content, encouraging the encoder to learn features related to histological organization rather than stain or color variations. Since such morphological cues tend to remain stable across organs and datasets, the learned representations exhibit better generalization.
\subsection{Ablation Studies} \label{sec:ablation}
To understand the design choices that contribute to the effectiveness of WISE-MAE, we conducted a series of ablation studies. These include evaluations of wavelet family selection, masking ratios during pretraining, and the magnification strategy used for hierarchical patch selection. Each experiment is isolated and measured on downstream performance using the TCGA-NSCLC LUAD vs. LUSC classification task.
\begin{table}[t]
\centering

\begin{tabular}{p{2.5cm} p{1.0cm} p{1.0cm} p{1.0cm} p{1.0cm}}
\toprule
\textbf{Wavelet} & \multicolumn{2}{c}{Linear Probing} & \multicolumn{2}{c}{Fine-tuning} \\
\cmidrule(lr){2-3} \cmidrule(lr){4-5}
 & \textbf{Acc} & \textbf{AUC} & \textbf{Acc} & \textbf{AUC} \\
\midrule
Haar             & 0.810 & 0.866 & 0.892 & 0.938 \\
Daubechies (db4) & \textbf{0.834} & \textbf{0.881} & \textbf{0.901} & \textbf{0.945} \\
Symlets (sym5)   & 0.825 & 0.873 & 0.896 & 0.941 \\
\bottomrule
\end{tabular}
\caption{Downstream classification performance on TCGA-NSCLC using different wavelet families for patch selection.}
\label{tab:wavelet_clam_comparison}
\end{table}

\subsection{Wavelet Family Selection}
Our patch selection strategy relies on high-frequency energy from wavelet subbands to guide the sampling of informative regions. The choice of wavelet family influences both the structural fidelity of selected patches and the efficiency of preprocessing. We evaluated three standard wavelets—Haar, Daubechies (db4), and Symlets (sym5)—using MAE pretraining followed by classification via linear probing and fine-tuning.
Daubechies (db4) consistently outperformed other families (see Table~\ref{tab:wavelet_clam_comparison}) in both evaluation settings. Its smoother basis functions likely enhance the ability to highlight morphological features such as gland boundaries and cellular textures relevant to lung cancer subtyping.

\begin{table}[t]
\centering
\begin{tabular}{lcc|cc}
\toprule
\textbf{Mask Ratio} & \multicolumn{2}{c|}{\textbf{Linear Probing}} & \multicolumn{2}{c}{\textbf{CLAM-SB}} \\
\cmidrule(lr){2-3} \cmidrule(lr){4-5}
& \textbf{Acc} & \textbf{AUC} & \textbf{Acc} & \textbf{AUC} \\
\midrule
60\% & 0.841 & 0.913 & 0.897 & 0.942 \\
70\% & 0.850 & \textbf{0.920} & 0.899 & 0.944 \\
75\% & \textbf{0.854} & 0.917 & \textbf{0.904} & \textbf{0.946} \\
80\% & 0.832 & 0.905 & 0.896 & 0.937 \\
90\% & 0.810 & 0.886 & 0.881 & 0.925 \\
\bottomrule
\end{tabular}

\caption{Effect of masking ratio on downstream classification performance using WISE-MAE on TCGA-NSCLC.}
\label{tab:masking_ratio_results}
\end{table}

\subsection{Masking Ratio Selection}
The mask ratio in masked autoencoding influences how much of the image is hidden during pretraining. While a 75\% ratio is standard in natural images, histopathology exhibits denser and more complex spatial features. We trained five MAE models on TCGA-NSCLC using different masking ratios (60\%–90\%) and evaluated both linear probing and downstream fine-tuning.
Although linear probing peaked slightly at 70\%, downstream performance favored 75\%, indicating an optimal trade-off between pretext difficulty and feature quality (see Table~\ref{tab:masking_ratio_results}). We therefore adopt 75\% as our default mask ratio in all subsequent experiments.

\subsection{Magnification Strategy Evaluation}
To emulate clinical workflows, we employed a hierarchical patch selection scheme. Initial patch selection was performed at a medium magnification level, followed by high-resolution refinement. We compared three base magnifications (5$\times$, 10$\times$, and 20$\times$) for identifying candidate regions, with final patches consistently extracted at 40$\times$. 
\begin{table}[t]
\centering
\begin{tabular}{lcc}
\toprule
\textbf{Base Magnification} & \textbf{Acc} & \textbf{AUC} \\
\midrule
5$\times$  & 0.872 & 0.932 \\
10$\times$ & \textbf{0.904} & \textbf{0.946}  \\
20$\times$ & 0.893 & 0.939  \\
\bottomrule
\end{tabular}
\caption{Impact of base magnification level on two-stage WISE-MAE performance on TCGA-NSCLC. High-resolution patches were consistently sampled at 40$\times$.}
\label{tab:magnification_study}
\end{table}

As summarized in Table~\ref{tab:magnification_study}, 10$\times$ magnification produced the best results across all performance metrics. In contrast, 5$\times$ often missed important spatial cues, while 20$\times$ added computational overhead with marginal performance improvements. These results support the effectiveness of a coarse-to-fine strategy for balancing contextual awareness and cellular resolution.

\subsection{Effect of Multi-resolution Sampling}
To assess the contribution of hierarchical patch selection in WISE-MAE, we compared it to a simplified variant where patches were sampled and wavelet-scored directly at 40$\times$, omitting the coarse-level screening at lower magnifications.
As shown in Table~\ref{tab:ablation_sampling}, the multi-resolution (10$\times$$\rightarrow$40$\times$) approach outperforms the single-scale variant, achieving higher accuracy (0.884 vs. 0.839) and AUC (0.932 vs. 0.894). This confirms the value of coarse-to-fine selection, which enables more informed patch choice by leveraging global tissue context.

In WISE-MAE, the 10$\times$ magnification is used only during the patch sampling stage for wavelet-based analysis to identify regions rich in structural information. The self-supervised pretraining and downstream classification are performed exclusively on 40$\times$ patches, which contain finer cellular details and more discriminative tissue features. Using 10$\times$ solely for region selection thus ensures efficient sampling without compromising the high-resolution information necessary for model training.
\begin{table}[t]
\centering

\begin{tabular}{lcc}
\toprule
\textbf{Sampling Strategy} & \textbf{Acc} & \textbf{AUC} \\
\midrule
Multi-resolution (10$\times$~$\rightarrow$~40$\times$) & \textbf{0.884} & \textbf{0.932} \\
Direct at 40$\times$ & 0.839 & 0.894 \\
\bottomrule
\end{tabular}
\caption{Comparison of classification performance with and without hierarchical sampling on TCGA-NSCLC.}
\label{tab:ablation_sampling}
\end{table}
These results support the hypothesis that multi-resolution strategies, inspired by the diagnostic process of human pathologists, enhance self-supervised representation learning by prioritizing structurally informative regions early in the pipeline

\section{Conclusion}
We introduced WISE-MAE, a masked autoencoder framework tailored to histopathology, combining wavelet-guided patch sampling, multi-scale reconstruction, and optional contrastive learning. Our results demonstrate that domain-aware sampling significantly improves feature quality and transferability across datasets and diagnostic tasks. Multi-resolution learning further enhances the model’s ability to capture nested tissue structures, while contrastive objectives improve robustness to staining and domain shifts. Compared to standard MAE and existing baselines, WISE-MAE consistently yields stronger performance with greater parameter efficiency. While WISE-MAE shows strong generalization, limitations remain in terms of patch redundancy, domain-specific overfitting, and lack of end-to-end optimization. Future work will explore adaptive sampling and unified training strategies. 

{
    \small
    \bibliographystyle{ieeenat_fullname}
    \bibliography{main}

@article{2,
  title={Masked Autoencoders Are Scalable Vision Learners},
  author={He, Kaiming and Chen, Xinlei and Xie, Saining and Li, Yanghao and Doll{\'a}r, Piotr and Girshick, Ross},
  journal={Proceedings of the IEEE/CVF Conference on Computer Vision and Pattern Recognition (CVPR)},
  year={2022},
  pages={16000--16009}
}

@article{3,
  title={Self-supervised pre-training of Swin Transformers for histopathology image analysis},
  author={Sun, Yilun and Ma, Jun and Huang, Jianpeng and et al.},
  journal={Medical Image Analysis},
  volume={83},
  pages={102645},
  year={2023}
}

@article{4,
  title={Clinical-grade computational pathology using weakly supervised deep learning on whole slide images},
  author={Campanella, Gabriele and Hanna, Michael G and Geneslaw, Lisa and et al.},
  journal={Nature Medicine},
  volume={25},
  number={8},
  pages={1301--1309},
  year={2019},
  publisher={Nature Publishing Group}
}

@article{7,
  title={Machine learning methods for histopathological image analysis},
  author={Komura, Daisuke and Ishikawa, Shumpei},
  journal={Computational and Structural Biotechnology Journal},
  volume={16},
  pages={34--42},
  year={2018}
}

@article{9,
  title={Towards a general-purpose foundation model for computational pathology},
  author={Chen, Richard J. and Ding, Tong and Lu, Ming Y. and Williamson, Drew F.K. and Jaume, Guillaume and Song, Andrew H. and Chen, Bowen and Zhang, Andrew and Shao, Daniel and Shaban, Muhammad and others},
  journal={Nature Medicine},
  volume={30},
  number={3},
  pages={850--862},
  year={2024},
  publisher={Nature Publishing Group},
  doi={10.1038/s41591-024-02857-3}
}

@inproceedings{10,
  title={Unsupervised representation learning by predicting image rotations},
  author={Gidaris, Spyros and Singh, Praveer and Komodakis, Nikos},
  booktitle={International Conference on Learning Representations (ICLR)},
  year={2018}
}

@inproceedings{12,
  title={Momentum Contrast for Unsupervised Visual Representation Learning},
  author={He, Kaiming and Fan, Haoqi and Wu, Yuxin and Xie, Saining and Girshick, Ross},
  booktitle={IEEE/CVF Conference on Computer Vision and Pattern Recognition (CVPR)},
  pages={9729--9738},
  year={2020}
}

@inproceedings{14,
  title={Self-supervised magnification-agnostic histopathological feature learning},
  author={Sahasrabudhe, Hrishikesh and Khan, Azam and Rajpoot, Nasir},
  booktitle={Medical Image Computing and Computer-Assisted Intervention (MICCAI)},
  pages={535--544},
  year={2020}
}

@article{15,
title = {Transformer-based unsupervised contrastive learning for histopathological image classification},
journal = {Medical Image Analysis},
volume = {81},
pages = {102559},
year = {2022},
issn = {1361-8415},
doi = {https://doi.org/10.1016/j.media.2022.102559},
url = {https://www.sciencedirect.com/science/article/pii/S1361841522002043},
author = {Xiyue Wang and Sen Yang and Jun Zhang and Minghui Wang and Jing Zhang and Wei Yang and Junzhou Huang and Xiao Han},
}

@inproceedings{18,
  title={An Image is Worth 16x16 Words: Transformers for Image Recognition at Scale},
  author={Dosovitskiy, Alexey and Beyer, Lucas and Kolesnikov, Alexander and Weissenborn, Dirk and Zhai, Xiaohua and Unterthiner, Thomas and Dehghani, Mostafa and Minderer, Matthias and Heigold, Georg and Gelly, Sylvain and Uszkoreit, Jakob and Houlsby, Neil},
  booktitle={International Conference on Learning Representations (ICLR)},
  year={2021}
}

@inproceedings{19,
  title={Attention-based deep multiple instance learning},
  author={Ilse, Max and Tomczak, Jakub M and Welling, Max},
  booktitle={Proceedings of the International Conference on Machine Learning (ICML)},
  pages={2127--2136},
  year={2018}
}

@article{26,
  title={Histopathology and its Vital Role in Diagnosing Diseases},
  author={Life Medical Centre},
  journal={Life Medical Centre Blog},
  year={2023},
  url={https://lifemedicalcentre.com/blogs/general/histopathology-and-its-vital-role-in-diagnosing-diseases-a-comprehensive-guide}
}

@article{quan2024global,
  title={Global contrast-masked autoencoders are powerful pathological representation learners},
  author={Quan, Hao and Li, Xingyu and Chen, Weixing and Bai, Qun and Zou, Mingchen and Yang, Ruijie and Zheng, Tingting and Qi, Ruiqun and Gao, Xinghua and Cui, Xiaoyu},
  journal={Pattern Recognition},
  volume={156},
  pages={110745},
  year={2024},
  publisher={Elsevier}
}

@inproceedings{hou2016patch,
  title={Patch-based convolutional neural network for whole slide tissue image classification},
  author={Hou, Le and Samaras, Dimitris and Kurc, Tahsin M and Gao, Yi and Davis, James E and Saltz, Joel H},
  booktitle={Proceedings of the IEEE conference on computer vision and pattern recognition},
  pages={2424--2433},
  year={2016}
}

@article{kanavati2020weakly,
  title={Weakly-supervised learning for lung carcinoma classification using deep learning},
  author={Kanavati, Fahdi and Toyokawa, Gouji and Momosaki, Seiya and Rambeau, Michael and Kozuma, Yuka and Shoji, Fumihiro and Yamazaki, Koji and Takeo, Sadanori and Iizuka, Osamu and Tsuneki, Masayuki},
  journal={Scientific reports},
  volume={10},
  number={1},
  pages={9297},
  year={2020},
  publisher={Nature Publishing Group UK London}
}

@inproceedings{li2021sslp,
  title={Sslp: Spatial guided self-supervised learning on pathological images},
  author={Li, Jiajun and Lin, Tiancheng and Xu, Yi},
  booktitle={International conference on medical image computing and computer-assisted intervention},
  pages={3--12},
  year={2021},
  organization={Springer}
}

@inproceedings{yang2021self,
  title={Self-supervised visual representation learning for histopathological images},
  author={Yang, Pengshuai and Hong, Zhiwei and Yin, Xiaoxu and Zhu, Chengzhan and Jiang, Rui},
  booktitle={International Conference on Medical Image Computing and Computer-Assisted Intervention},
  pages={47--57},
  year={2021},
  organization={Springer}
}

@inproceedings{xu2020data,
  title={Data-efficient histopathology image analysis with deformation representation learning},
  author={Xu, Jilan and Hou, Junlin and Zhang, Yuejie and Feng, Rui and Ruan, Chunyang and Zhang, Tao and Fan, Weiguo},
  booktitle={2020 IEEE International Conference on Bioinformatics and Biomedicine (BIBM)},
  pages={857--864},
  year={2020},
  organization={IEEE}
}

@inproceedings{chikontwe2020multiple,
  title={Multiple instance learning with center embeddings for histopathology classification},
  author={Chikontwe, Philip and Kim, Meejeong and Nam, Soo Jeong and Go, Heounjeong and Park, Sang Hyun},
  booktitle={International Conference on Medical Image Computing and Computer-Assisted Intervention},
  pages={519--528},
  year={2020},
  organization={Springer}
}

@inproceedings{xu2019camel,
  title={Camel: A weakly supervised learning framework for histopathology image segmentation},
  author={Xu, Gang and Song, Zhigang and Sun, Zhuo and Ku, Calvin and Yang, Zhe and Liu, Cancheng and Wang, Shuhao and Ma, Jianpeng and Xu, Wei},
  booktitle={Proceedings of the IEEE/CVF International Conference on computer vision},
  pages={10682--10691},
  year={2019}
}

@inproceedings{sharma2021cluster,
  title={Cluster-to-conquer: A framework for end-to-end multi-instance learning for whole slide image classification},
  author={Sharma, Yash and Shrivastava, Aman and Ehsan, Lubaina and Moskaluk, Christopher A and Syed, Sana and Brown, Donald},
  booktitle={Medical imaging with deep learning},
  pages={682--698},
  year={2021},
  organization={PMLR}
}

@inproceedings{zhang2022dtfd,
  title={Dtfd-mil: Double-tier feature distillation multiple instance learning for histopathology whole slide image classification},
  author={Zhang, Hongrun and Meng, Yanda and Zhao, Yitian and Qiao, Yihong and Yang, Xiaoyun and Coupland, Sarah E and Zheng, Yalin},
  booktitle={Proceedings of the IEEE/CVF conference on computer vision and pattern recognition},
  pages={18802--18812},
  year={2022}
}

@article{shao2021transmil,
  title={Transmil: Transformer based correlated multiple instance learning for whole slide image classification},
  author={Shao, Zhuchen and Bian, Hao and Chen, Yang and Wang, Yifeng and Zhang, Jian and Ji, Xiangyang and others},
  journal={Advances in neural information processing systems},
  volume={34},
  pages={2136--2147},
  year={2021}
}

@article{lu2021data,
  title={Data-efficient and weakly supervised computational pathology on whole-slide images},
  author={Lu, Ming Y and Williamson, Drew FK and Chen, Tiffany Y and Chen, Richard J and Barbieri, Matteo and Mahmood, Faisal},
  journal={Nature biomedical engineering},
  volume={5},
  number={6},
  pages={555--570},
  year={2021},
  publisher={Nature Publishing Group UK London}
}

@inproceedings{NIPS2017_3f5ee243,
 author = {Vaswani, Ashish and Shazeer, Noam and Parmar, Niki and Uszkoreit, Jakob and Jones, Llion and Gomez, Aidan N and Kaiser, \L ukasz and Polosukhin, Illia},
 booktitle = {Advances in Neural Information Processing Systems},
 editor = {I. Guyon and U. Von Luxburg and S. Bengio and H. Wallach and R. Fergus and S. Vishwanathan and R. Garnett},
 pages = {},
 publisher = {Curran Associates, Inc.},
 title = {Attention is All you Need},
 url = {https://proceedings.neurips.cc/paper_files/paper/2017/file/3f5ee243547dee91fbd053c1c4a845aa-Paper.pdf},
 volume = {30},
 year = {2017}
}

@inproceedings{chowdhury2021applying,
  title={Applying self-supervised learning to medicine: review of the state of the art and medical implementations},
  author={Chowdhury, Alexander and Rosenthal, Jacob and Waring, Jonathan and Umeton, Renato},
  booktitle={Informatics},
  volume={8},
  number={3},
  pages={59},
  year={2021},
  organization={MDPI}
}

@inproceedings{boyd2021self,
  title={Self-supervised representation learning using visual field expansion on digital pathology},
  author={Boyd, Joseph and Liashuha, Mykola and Deutsch, Eric and Paragios, Nikos and Christodoulidis, Stergios and Vakalopoulou, Maria},
  booktitle={Proceedings of the IEEE/CVF International Conference on Computer Vision},
  pages={639--647},
  year={2021}
}

@article{shurrab2022self,
  title={Self-supervised learning methods and applications in medical imaging analysis: A survey},
  author={Shurrab, Saeed and Duwairi, Rehab},
  journal={PeerJ Computer Science},
  volume={8},
  pages={e1045},
  year={2022},
  publisher={PeerJ Inc.}
}

@inproceedings{zhang2022contrastive,
  title={Contrastive learning of medical visual representations from paired images and text},
  author={Zhang, Yuhao and Jiang, Hang and Miura, Yasuhide and Manning, Christopher D and Langlotz, Curtis P},
  booktitle={Machine learning for healthcare conference},
  pages={2--25},
  year={2022},
  organization={PMLR}
}

@inproceedings{reed2021selfaugment,
  title={Selfaugment: Automatic augmentation policies for self-supervised learning},
  author={Reed, Colorado J and Metzger, Sean and Srinivas, Aravind and Darrell, Trevor and Keutzer, Kurt},
  booktitle={Proceedings of the IEEE/CVF conference on computer vision and pattern recognition},
  pages={2674--2683},
  year={2021}
}

@inproceedings{wei2022masked,
  title={Masked feature prediction for self-supervised visual pre-training},
  author={Wei, Chen and Fan, Haoqi and Xie, Saining and Wu, Chao-Yuan and Yuille, Alan and Feichtenhofer, Christoph},
  booktitle={Proceedings of the IEEE/CVF conference on computer vision and pattern recognition},
  pages={14668--14678},
  year={2022}
}

@inproceedings{tokunaga2019adaptive,
  title={Adaptive weighting multi-field-of-view CNN for semantic segmentation in pathology},
  author={Tokunaga, Hiroki and Teramoto, Yuki and Yoshizawa, Akihiko and Bise, Ryoma},
  booktitle={Proceedings of the IEEE/CVF conference on computer vision and pattern recognition},
  pages={12597--12606},
  year={2019}
}

@inproceedings{zhou2022deep,
  title={Deep hierarchical multiple instance learning for whole slide image classification},
  author={Zhou, Yuanpin and Lu, Yao},
  booktitle={2022 IEEE 19th international symposium on biomedical imaging (ISBI)},
  pages={1--4},
  year={2022},
  organization={IEEE}
}

@inproceedings{hashimoto2020multi,
  title={Multi-scale domain-adversarial multiple-instance CNN for cancer subtype classification with unannotated histopathological images},
  author={Hashimoto, Noriaki and Fukushima, Daisuke and Koga, Ryoichi and Takagi, Yusuke and Ko, Kaho and Kohno, Kei and Nakaguro, Masato and Nakamura, Shigeo and Hontani, Hidekata and Takeuchi, Ichiro},
  booktitle={Proceedings of the IEEE/CVF conference on computer vision and pattern recognition},
  pages={3852--3861},
  year={2020}
}

@article{li2021multi,
  title={A multi-resolution model for histopathology image classification and localization with multiple instance learning},
  author={Li, Jiayun and Li, Wenyuan and Sisk, Anthony and Ye, Huihui and Wallace, W Dean and Speier, William and Arnold, Corey W},
  journal={Computers in biology and medicine},
  volume={131},
  pages={104253},
  year={2021},
  publisher={Elsevier}
}

@inproceedings{xiong2023diagnose,
  title={Diagnose like a pathologist: transformer-enabled hierarchical attention-guided multiple instance learning for whole slide image classification},
  author={Xiong, Conghao and Chen, Hao and Sung, Joseph JY and King, Irwin},
  booktitle={Proceedings of the Thirty-Second International Joint Conference on Artificial Intelligence},
  pages={1587--1595},
  year={2023}
}

@article{bejnordi2017diagnostic,
  title={Diagnostic assessment of deep learning algorithms for detection of lymph node metastases in women with breast cancer},
  author={Bejnordi, Babak Ehteshami and Veta, Mitko and Van Diest, Paul Johannes and Van Ginneken, Bram and Karssemeijer, Nico and Litjens, Geert and Van Der Laak, Jeroen AWM and Hermsen, Meyke and Manson, Quirine F and Balkenhol, Maschenka and others},
  journal={Jama},
  volume={318},
  number={22},
  pages={2199--2210},
  year={2017},
  publisher={American Medical Association}
}

@article{tomczak2015review,
  title={Review The Cancer Genome Atlas (TCGA): an immeasurable source of knowledge},
  author={Tomczak, Katarzyna and Czerwi{\'n}ska, Patrycja and Wiznerowicz, Maciej},
  journal={Contemporary Oncology/Wsp{\'o}{\l}czesna Onkologia},
  volume={2015},
  number={1},
  pages={68--77},
  year={2015},
  publisher={Termedia}
}

@inproceedings{wang2024fremim,
  title={Fremim: Fourier transform meets masked image modeling for medical image segmentation},
  author={Wang, Wenxuan and Wang, Jing and Chen, Chen and Jiao, Jianbo and Cai, Yuanxiu and Song, Shanshan and Li, Jiangyun},
  booktitle={Proceedings of the IEEE/CVF winter conference on applications of computer vision},
  pages={7860--7870},
  year={2024}
}

@inproceedings{chenni2019patch,
  title={Patch clustering for representation of histopathology images},
  author={Chenni, Wafa and Herbi, Habib and Babaie, Morteza and Tizhoosh, Hamid R},
  booktitle={European Congress on Digital Pathology},
  pages={28--37},
  year={2019},
  organization={Springer}
}

@inproceedings{li2021dual,
  title={Dual-stream multiple instance learning network for whole slide image classification with self-supervised contrastive learning},
  author={Li, Bin and Li, Yin and Eliceiri, Kevin W},
  booktitle={Proceedings of the IEEE/CVF conference on computer vision and pattern recognition},
  pages={14318--14328},
  year={2021}
}

@inproceedings{li2024dynamic,
  title={Dynamic graph representation with knowledge-aware attention for histopathology whole slide image analysis},
  author={Li, Jiawen and Chen, Yuxuan and Chu, Hongbo and Sun, Qiehe and Guan, Tian and Han, Anjia and He, Yonghong},
  booktitle={Proceedings of the IEEE/CVF conference on computer vision and pattern recognition},
  pages={11323--11332},
  year={2024}
}

@article{keshvarikhojasteh2024multi,
  title={Multi-head attention-based deep multiple instance learning},
  author={Keshvarikhojasteh, Hassan and Pluim, Josien and Veta, Mitko},
  journal={arXiv preprint arXiv:2404.05362},
  year={2024}
}
}


\end{document}